# Bandit Algorithms for Tree Search

**Pierre-Arnaud Coquelin**
CMAP, Ecole Polytechnique
91128 Palaiseau Cedex, France
coquelin@cmapx.polytechnique.fr

**Rémi Munos**
SequeL project, INRIA Futurs Lille
40 avenue Halley,
59650 Villeneuve d'Ascq, France
remi.munos@inria.fr

## Abstract

Bandit based methods for tree search have recently gained popularity when applied to huge trees, e.g. in the game of go [6]. Their efficient exploration of the tree enables to return rapidly a good value, and improve precision if more time is provided. The *UCT algorithm* [8], a tree search method based on Upper Confidence Bounds (UCB) [2], is believed to adapt locally to the effective smoothness of the tree. However, we show that UCT is "over-optimistic" in some sense, leading to a worst-case regret that may be very poor. We propose alternative bandit algorithms for tree search. First, a modification of UCT using a confidence sequence that scales exponentially in the horizon depth is analyzed. We then consider *Flat-UCB* performed on the leaves and provide a finite regret bound with high probability. Then, we introduce and analyze a *Bandit Algorithm for Smooth Trees* (BAST) which takes into account actual smoothness of the rewards for performing efficient "cuts" of sub-optimal branches with high confidence. Finally, we present an incremental tree expansion which applies when the full tree is too big (possibly infinite) to be entirely represented and show that with high probability, only the optimal branches are indefinitely developed. We illustrate these methods on a global optimization problem of a continuous function, given noisy values.

## 1 Introduction

Bandit algorithms have been used recently for tree search, because of their efficient trading-off between exploration of the most uncertain branches and exploitation of the most promising ones, leading to very promising results for dealing with huge trees (see e.g. the go program MoGo in [6]). In this paper we focus on bandit algorithms based on *Upper Confidence Bounds* (UCB) [2] applied to tree search, such as UCT (Upper Confidence Bounds applied to Trees) [8]. This general bandit-based procedure for tree search is defined by Algorithm 1; the core issue being the way the upper-bounds $B_{i,p,n_i}$ on the value of each node $i$ are maintained.

---
**Algorithm 1** Bandit Algorithm for Tree Search

　**for** $n \geq 1$ **do**
　　**Run the $n$-th trajectory (sequence of nodes $(i_0, \ldots, i_D)$ from the root to a leaf):**
　　Set the current node $i_0$ to the root
　　**for** $d = 1$ **to** $D$ **do**
　　　* Compute the bounds $B_{j,n_{i_{d-1}},n_j}$ for all children $j \in \mathcal{C}(i_{d-1})$ of node $i_{d-1}$
　　　* Select the node $i_d$ as a child of node $i_{d-1}$ that has the highest $B$ value, i.e.
$$i_d \in \arg\max_{j \in \mathcal{C}(i_{d-1})} B_{j,n_{i_{d-1}},n_j}$$
　　**end for**
　　Receive reward $x_n \stackrel{iid}{\sim} X_{i_D}$
　**end for**

---

A trajectory is a sequence of nodes from the root to a leaf, where at each node $i$, the next node is chosen as the (or one) child having the highest $B$ value among its children $\mathcal{C}(i)$. A reward is received at the leaf. After a trajectory is run, the number of visits $n_i$ of each node $i$ in the trajectory are incremented.

For example, in the case of UCT, the upper-confidence bound of a node $i$ is:

$$B_{i,p,n_i} \stackrel{\text{def}}{=} X_{i,n_i} + \sqrt{\frac{2\log(p)}{n_i}}. \qquad (1)$$

where $n_i$ (resp. $p$) is the number of times this node (resp. its parent) has been visited, and $X_{i,n_i}$ is the empirical mean of the rewards that have been obtained by trajectories going through that node.

In this paper we consider a max search in a binary tree (i.e. there are 2 actions from each node) of depth $D$. The extension to more actions is straightforward. At each leaf $i$ is assigned a random variable $X_i$, with bounded support included in $[0, 1]$, whose law is unknown.

We write $I_t$ the chosen leaf at round $t$ (i.e. obtained at the end of the $t$−th trajectory). Successive visits of the leaves $(I_t)$ yield a sequence of independent and identically distributed (i.i.d.) samples $x_t \sim X_{I_t}$, called *rewards*, or *payoffs*. We write $x_{i,t}$ the $t$-th reward received at a leaf $i$.

The value of a leaf $i$ is the expected reward: $\mu_i \stackrel{\text{def}}{=} \mathbb{E}X_i$. Now we define the value of any node $i$ as the maximal value of the leaves in the sub-tree (branch) starting from node $i$. An optimal leaf is a leaf having the largest expected reward. We will denote by $*$ quantities related to an optimal node. For example $\mu^*$ denote the maximal value of the leaves, i.e. the value of the root. One possible goal is to compute $\mu^*$. Another goal is to find an optimal branch, i.e. a sequence of nodes from the root to an optimal leaf. Let $i^*$ be an optimal leaf. We write $n_i(t)$ the random variable that counts the number of times a node $i$ has been visited up to time $t$. Thus $x_t = x_{I_t, n_{I_t}(t)}$.

We define the *cumulative regret* up to time $n$ as the loss in the cumulative rewards resulting from choosing sub-optimal leaves instead of an optimal one:

$$R_n \stackrel{\text{def}}{=} \sum_{t=1}^{n} x_{i^*, n_{i^*}(t)} - \sum_{t=1}^{n} x_t.$$

We also define the *cumulative pseudo-regret*:

$$\bar{R}_n \stackrel{\text{def}}{=} n\mu^* - \sum_{t=1}^{n} \mu_{I_t} = \sum_{j \in \mathcal{L}} n_j \Delta_j,$$

where $\mathcal{L}$ is the set of leaves and $\Delta_j \stackrel{\text{def}}{=} \mu^* - \mu_j$. The difference between the regret and the pseudo-regret comes from the randomness of the rewards.

In tree search, our goal is thus to find an exploration policy of the branches such as to minimize the regret, in order to select an optimal leaf as fast as possible. Notice that thanks to Wald's theorem, the regret and pseudo-regret have the same expectation: $\mathbb{E}R_n = \mathbb{E}\bar{R}_n$. Now, thanks to a contraction of measure phenomenon, the regret per round $R_n/n$ turns out to be very close to the pseudo-regret per round $\bar{R}_n/n$. Indeed, using Azuma's inequality for martingale difference sequences (see Proposition 1), with probability at least $1 - \beta$, we have at time $n$,

$$|R_n - \bar{R}_n| \le \sqrt{|\text{Sub}(n)| \log(2/\beta)/2}, \quad (2)$$

where $|\text{Sub}(n)|$ is the cardinal of $\text{Sub}(n) \stackrel{\text{def}}{=} \{t \in \{1, \ldots, n\}, I_t \ne i^*\}$, the set of times $t$, up to time $n$, when the chosen leaf $I_t$ is different from $i^*$.

The fact that $R_n - \bar{R}_n$ is a martingale difference sequence comes from the property that the chosen leaf at time $t$ is entirely determined by the filtration $\mathcal{F}_{t-1}$ (defined by the random samples up to time $t-1$). Thus $\mathbb{E}[x_t | \mathcal{F}_{t-1}] = \mu_{I_t}$ and $\bar{R}_n - R_n = \sum_{t \in \text{Sub}(n)} x_t - \mu_{I_t}$ with $\mathbb{E}[x_t - \mu_{I_t} | \mathcal{F}_{t-1}] = 0$, and (2) follows from Proposition 1. Hence, in this paper, we will focus on providing high probability bounds on the pseudo-regret, and on the number of sampled sub-optimal leaves. Bounds on the regret directly follow from (2).

The paper is organized as follows. First, we analyze the UCT algorithm defined by the upper confidence bound (1). We show that its behavior is risky and may lead to a regret (expressed in terms of the depth $D$ of the tree) as bad as $\Omega(\exp(\cdots \exp(1) \cdots))$ (there are $D - 1$ composed exponential functions). If we modify the algorithm by using an increased exploration sequence, defining:

$$B_{i,p,n_i} \stackrel{\text{def}}{=} X_{i,n_i} + \sqrt{\frac{\sqrt{p}}{n_i}}, \quad (3)$$

we obtain an improved worst-case behavior compared to regular UCT, but the regret may still be as bad as $\Omega(\exp(\exp(D)))$ (see Section 2). We then propose in Section 3 a *modified UCT* based on the bound (3), where the confidence interval is multiplied by a factor that scales exponentially with the horizon depth.

Next, we analyze the *Flat-UCB* algorithm, which simply performs UCB directly on the leaves.

In Section 5 we introduce the **Bandit Algorithm for Smooth Trees** (BAST), which takes into account actual smoothness of the rewards for performing efficient "cuts" of branches that are (with high confidence) sub-optimal, without having to explore all the leaves of these branches. This is our main contribution. We analyze the regret and show the improvement over both UCT and Flat-UCB. We provide a numerical experiment for the problem of optimizing a continuous function given noisy values.

Finally, in Section 6 we extend and analyze BAST using an incremental tree expansion procedure, which applies when the tree is so big that its full representation in memory is impossible. We show that this method builds an asymmetric tree that indefinitely develops the optimal branch only.

**Preliminaries and additional notations** Let $\mathcal{L}$ denotes the set of leaves and $\mathcal{S}$ the set of sub-optimal leaves. For any node $i$, we write $\mathcal{L}(i)$ the set of leaves that belong to the sub-tree starting from node $i$. When there is no possible confusion, we write $n_i$ instead of $n_i(n)$ the number of times a node $i$ has been visited up to time $n$. For a leaf $j$, we define the empirical mean of the rewards obtained at $j$:

$$X_{j,n_j} \stackrel{\text{def}}{=} \frac{1}{n_j} \sum_{t=1}^{n_j} x_{j,t}.$$

Now, for any node $i$, we defined the empirical mean of the rewards obtained from $i$, up to time $n$:

$$X_{i,n_i} \stackrel{\text{def}}{=} \frac{1}{n_i} \sum_{t=1}^{n_i} x_{i,t} = \frac{1}{n_i} \sum_{j \in \mathcal{L}(i)} n_j X_{j,n_j},$$

and the empirical mean of the values of the leaves:

$$\bar{X}_{i,n_i} \stackrel{\text{def}}{=} \frac{1}{n_i} \sum_{j \in \mathcal{L}(i)} n_j \mu_j.$$

We finally remind Azuma's inequality (see [7]):

**Proposition 1.** *Let $Y_1, Y_2, \cdots$ be a martingale difference sequence, i.e. for all $t \geq 1$, $\mathbb{E}[Y_t|Y_1, \ldots, Y_{t-1}] = 0$ with probability (w.p.) 1, and $d_1, d_2, \ldots$ real positive numbers such that for all $t \geq 1$, $0 \leq Y_i \leq d_i$ w.p. 1. Then for every $\epsilon > 0$,*

$$\mathbb{P}(|\sum_{t=1}^{n} Y_t| \geq \epsilon) \leq 2e^{-2\epsilon^2/\sum_{t=1}^{n} d_t^2}.$$

## 2 Regret lower bound for UCT

Notice that the bound (1) used in UCT comes from a concentration inequality that applies when the rewards are independent and identically distributed. However, in the tree problem, the received rewards at each node do not satisfy these assumptions because the chosen leaves depend on a non-stationary node selection process. Hence, the bound (1) may not be a true upper confidence bound of the nodes value.

Nevertheless, as argued in [8], since the confidence interval term increases with $\log(p)$ when a child node is not chosen, all children from each node will eventually be indefinitely visited. Then, using an inductive argument on the depth, it is proven that, after a transitory period of time $N_0$, only the optimal branch will be followed, yielding an expected regret of order $O(N_0 + \log(n))$. However, this "transitory" phase may last very long.

Indeed, consider the example shown in Figure 1. The rewards are deterministic and for a node of depth $d$ in the optimal branch (obtained after choosing $d$ times action 1), if action 2 is chosen, then a reward of $\frac{D-d}{D}$ is received (all leaves in this branch have the same reward). If action 1 is chosen, then this moves to the next node in the optimal branch. At depth $D-1$, action 1 yields reward 1 and action 2, reward 0. We assume that when a node is visited for the first time, the algorithm starts by choosing action 2 before choosing action 1.

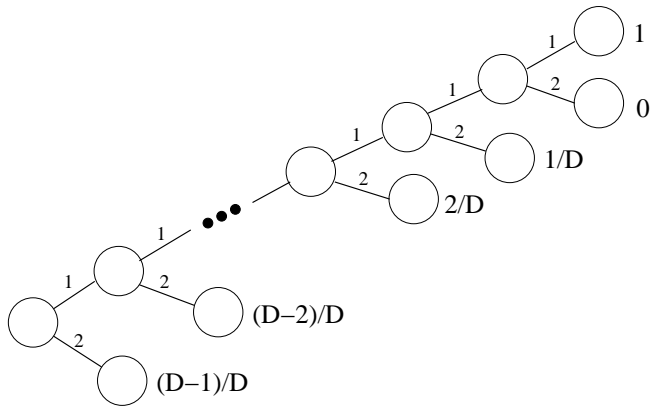

Figure 1: A bad example for UCT. From the root (left node), action 2 leads to a node from which all leaves yield reward $\frac{D-1}{D}$. The optimal branch consists in choosing always action 1, which yields reward 1. In the beginning, the algorithm believes the arm 2 is the best, spending most of its times exploring this branch (as well as all other sub-optimal branches). It takes $\Omega(\exp(\ldots(\exp(1))\ldots))$ ($D-1$ times) rounds to get the 1 reward!

We now establish a lower bound on the number of times suboptimal rewards are received before getting the optimal 1 reward for the first time. Write $n$ the first time the optimal leaf is reached. Write $n_d$ the number of times the node (also written $d$ making a slight abuse of notation) of depth $d$ in the optimal branch is reached. Thus $n = n_0$ and $n_D = 1$. At depth $D-1$, we have $n_{D-1} = 2$ (since action 2 has been chosen once in node $D-1$).

We consider both the logarithmic confidence sequence used in (1) and the square root sequence in (3). Let us start with the square root confidence sequence (3). At depth $d-1$, since the optimal branch is followed by the $n$-th trajectory, we have (writting $d'$ the node resulting from action 2 in the node $d-1$): $X_{d',n_{d'}} + \sqrt{\frac{\sqrt{n_{d-1}}}{n_{d'}}} \leq X_{d,n_d} + \sqrt{\frac{\sqrt{n_{d-1}}}{n_d}}$. But $X_{d',n_{d'}} = (D-d)/D$ and $X_{d,n_d} \leq (D-(d+1))/D$ since the 1 reward has not been received before. We deduce that $\frac{1}{D} \leq \sqrt{\frac{\sqrt{n_{d-1}}}{n_d}}$.

Thus for the square root confidence sequence, we have $n_{d-1} \geq n_d^2/D^4$. Now, by induction,

$$n \geq \frac{n_1^2}{D^4} \geq \frac{n_2^{2^2}}{D^{4(1+2)}} \geq \frac{n_3^{2^3}}{D^{4(1+2+3)}} \geq \cdots \geq \frac{n_{D-1}^{2^{D-1}}}{D^{2D(D-1)}}$$

Since $n_{D-1} = 2$, we obtain $n \geq \frac{2^{2^{D-1}}}{D^{2D(D-1)}}$. This is a double exponential dependency w.r.t. $D$. For example, for $D = 20$, we have $n \geq 10^{156837}$. Thus the transitory phase lasts $N_0 = \Omega(\exp(\exp(D)))$, and the regret is $\Omega(\exp(\exp(D))) + O(\log n)$, showing a double exponential dependency w.r.t. the depth of the tree.

Now, the usual logarithmic confidence sequence defined by (1) yields an even worst lower bound on the regret since we may show similarly that $n_{d-1} \geq \exp(n_d/(2D^2))$ thus $n = \Omega(\exp(\exp(\cdots \exp(1) \cdots)))$ (composition of $D - 1$ exponential functions). Thus UCT algorithm has a regret $\Omega(\exp(\exp(\cdots \exp(1) \cdots))) + O(\log(n))$, showing a hyperexponential dependency w.r.t. $D$.

The reason for this bad behavior is that the algorithm is too optimistic (it does not explore enough and may take a very long time before discovering good branches that looked initially bad) since the bounds (1) and (3) are not true upper bounds on the nodes value, at least during the transitory period which may last very long.

## 3 Modified UCT

We modify the confidence sequence in order to explore more the nodes that are closer to the root than the leaves, taking into account the fact that the time needed to decrease the bias $(\mu_i - \bar{X}_{i,n_i})$ at a node $i$ of depth $d$ increases with the depth horizon $(D - d)$. For such a node $i$ of depth $d$, we define the upper confidence bound:

$$B_{i,n_i} \stackrel{\text{def}}{=} X_{i,n_i} + (k_d + 1)\sqrt{\frac{\log(\beta_{n_i}^{-1})}{2n_i}} + \frac{k'_d}{n_i}, \quad (4)$$

where $\beta_n \stackrel{\text{def}}{=} \frac{\beta}{2Nn(n+1)}$ with $N \stackrel{\text{def}}{=} 2^{D+1} - 1$ the number of nodes in the tree, and the coefficients:

$$k_d \stackrel{\text{def}}{=} \frac{1 + \sqrt{2}}{\sqrt{2}}[(1 + \sqrt{2})^{D-d} - 1] \quad (5)$$
$$k'_d \stackrel{\text{def}}{=} (3^{D-d} - 1)/2$$

Notice that we used a simplified notation, writing $B_{i,n_i}$ instead of $B_{i,p,n_i}$ since the bound does not depend on the number of visits of the parent's node.

We now provide a high probability bound on the regret which is exponential in the depth $D$ and is square root dependent on $n$. The proof is given in [4].

**Theorem 1.** *Let $\beta > 0$. Consider Algorithm 1 with the upper confidence bound (4). Then, with probability at least $1 - \beta$, for all $n \geq 1$, the pseudo-regret is bounded by*

$$\bar{R}_n \leq \frac{1 + \sqrt{2}}{2}[(1 + \sqrt{2})^D - 1]\sqrt{\log(\beta_n^{-1})n} + \frac{3^D - 1}{2}$$

The confidence intervals used in (4) are actually chosen such that the $B$ values represent true upper confidence bounds on the nodes value, i.e. with high probability (at least $1 - \beta$), for all node $i$, for all $n_i \geq 1$, $\mu_i \leq B_{i,n_i}$. Thus this procedure is safe, which prevents us from having bad behaviors for which the regret could be disastrous, like in regular UCT. However, contrarily to regular UCT, in good cases, the procedure does not adapt to the effective smoothness in the tree. For example, at the root level, the confidence interval is $O(\exp(D)/\sqrt{n})$ which leads to sampling almost uniformly both actions for a time $O(\exp(D))$. Thus, if the tree were to contain 2 branches, one only with zeros, one only with ones, this smoothness would not be taken into account by this method, and the regret would still be $O(\exp(D)\sqrt{n})$. Thus modified UCT is less optimistic than regular UCT but safer in a worst-case scenario.

## 4 Flat UCB

A method that would combine both the safety of modified UCT and the adaptivity of regular UCT is to consider a regular UCB algorithm on the leaves. Such a *flat UCB* could naturally be implemented in the tree structure by defining the upper confidence bound of a non-leaf node as the maximal value of the children's bound:

$$B_{i,n_i} \stackrel{\text{def}}{=} \begin{cases} X_{i,n_i} + \sqrt{\frac{\log(\beta_{n_i}^{-1})}{2n_i}} & \text{if } i \text{ is a leaf,} \\ \max_{j \in \mathcal{C}(i)} B_{j,n_j} & \text{otherwise.} \end{cases} \quad (6)$$

where we use $\beta_n \stackrel{\text{def}}{=} \frac{\beta}{2^{D+1}n(n+1)}$. We deduce:

**Theorem 2.** *Consider the flat UCB defined by Algorithm 1 and (6). Then, with probability at least $1 - \beta$, the pseudo-regret is bounded by a constant independent of $n$:*

$$\bar{R}_n \leq 6 \sum_{i \in \mathcal{S}} \frac{1}{\Delta_i} \log(\frac{2^{D+2}}{\Delta_i^2 \beta}) \leq 6 \frac{2^D}{\Delta} \log(\frac{2^{D+2}}{\Delta^2 \beta}),$$

*where* $\Delta \stackrel{\text{def}}{=} \min_{i \in \mathcal{S}} \Delta_i$.

*Proof.* Consider the event $\mathcal{E}$ under which, for all leaves $i$, for all $n \geq 1$, we have $|X_{i,n} - \mu_i| \leq c_n$, with the confidence intervals $c_n = \sqrt{\frac{\log(\beta_n^{-1})}{2n}}$. Then the event $\mathcal{E}$ holds with probability at least $1 - \beta$. Indeed, from a union bound argument, there are at most $2^D$ Chernoff-Hoeffding's inequalities (one for each leaf) of the form:

$$\mathbb{P}\left(|X_{i,n} - \mu_i| \leq c_n, \forall n \geq 1\right) \geq 1 - \sum_{n \geq 1} 2\beta_n = 1 - \frac{\beta}{2^D}.$$

Under the event $\mathcal{E}$, we now provide a regret bound by bounding the number of times each sub-optimal leaf is visited. Let $i \in \mathcal{S}$ be a sub-optimal leaf. Write $*$ an optimal leaf. If, at some round $n$, the leaf $i$ is chosen, this means that $X_{*,n_*} + c_{n_*} \leq X_{i,n_i} + c_{n_i}$. Using the (lower and upper) confidence interval bounds for leaves $i$ and $*$, we deduce that $\mu^* \leq \mu_i + 2c_{n_i}$. Thus $\left(\frac{\Delta_i}{2}\right)^2 \leq \frac{\log(\beta_{n_i}^{-1})}{2n_i}$. Hence, for all $n \geq 1$, $n_i$ is bounded by the smallest integer $n_0$ such that $\frac{n_0}{\log(\beta_{n_0}^{-1})} > 2/\Delta_i^2$. An upper bound on $n_0$ may be found by analyzing the zero of the function $x \to x - w\log(2^{D+1}x(x+1)\beta^{-1})$, writing $w = 2/\Delta_i^2$. A rough bound on $n_0$ is $3w\log(w2^{D+1}\beta^{-1})$. We deduce that the number of times a leaf $i$ is chosen is at most

$$n_i \leq \frac{6}{\Delta_i^2}\log(\frac{2}{\Delta_i^2}2^{D+1}\beta^{-1}).$$

The bound on the regret follows immediately from the property that $\bar{R}_n = \sum_{i \in \mathcal{S}} n_i \Delta_i$. □

This algorithm is safe in the sense that with high probability, the bounds defined by (6) are true upper bounds on the value on the leaves. However, since there are $2^D$ leaves, the regret still depends exponentially on the depth $D$.

In the next section, we introduce another UCB-based algorithm that takes into account possible smoothness of the rewards to process efficient "cuts" of sub-optimal branches with high confidence.

## 5  Bandit Algorithm for Smooth Trees

We would like to exploit the fact that if the leaves of a branch have similar values, then a confidence interval on the value of that branch may be made much tighter than the maximal confidence interval of its leaves (as processed in the Flat UCB). Indeed, assume that from a node $i$, all leaves $j \in \mathcal{L}(i)$ in the branch $i$ have values $\mu_j$, such that $\mu_i - \mu_j \leq \delta$. Then,

$$\mu_i \leq \frac{1}{n_i}\sum_{j \in \mathcal{L}(i)} n_j(\mu_j + \delta) \leq X_{i,n_i} + \delta + \bar{X}_{i,n_i} - X_{i,n_i},$$

and thanks to Azuma's inequality, the term $\bar{X}_{i,n_i} - X_{i,n_i}$ is bounded with probability $1-\beta$ by a confidence interval $\sqrt{\frac{\log(2\beta^{-1})}{2n_i}}$ which depends only on $n_i$ (and not on $n_j$ for $j \in \mathcal{L}(i)$). We now make such a smoothness assumption either along **an optimal path only** or along **any almost optimal path**.

**Assumption $A^*$:** for any depth $d < D$, there exists $\delta_d > 0$, such that for (at least) one optimal node $i$ of depth $d$ (i.e. such that $\mu_i = \mu^*$), we have, for all leaves $j \in \mathcal{L}(i)$ in the branch $i$, $\mu^* - \mu_j \leq \delta_d$.

A stronger assumption requires the smoothness assumption to hold for any $\eta$-optimal node $i$ (where $\eta > 0$), i.e. when $i \in I_\eta \stackrel{\text{def}}{=} \{\text{node } i, \Delta_i \leq \eta\}$.

**Assumption $A_\eta$:** For any $i \in I_\eta$ of depth $d < D$, there exists $\delta_d > 0$, such that for all $j \in \mathcal{L}(i)$, $\mu_i - \mu_j \leq \delta_d$.

Typical choices of the smoothness coefficients $\delta_d$ are exponential $\delta_d \stackrel{\text{def}}{=} \delta\gamma^d$ (with $\delta > 0$ and $\gamma < 1$), polynomial $\delta_d \stackrel{\text{def}}{=} \delta d^\alpha$ (with $\alpha < 0$), or linear $\delta_d \stackrel{\text{def}}{=} \delta(D-d)$ (Lipschitz in the tree distance) sequences.

We define the **Bandit Algorithm for Smooth Trees** (BAST) by Algorithm 1 with the upper confidence bounds defined, for any leaf $i$, by $B_{i,n_i} \stackrel{\text{def}}{=} X_{i,n_i} + c_{n_i}$, and for any non-leaf node $i$ of depth $d$, by

$$B_{i,n_i} \stackrel{\text{def}}{=} \min\left\{\max_{j \in \mathcal{C}(i)} B_{j,n_j}, X_{i,n_i} + \delta_d + c_{n_i}\right\} \quad (7)$$

with the confidence interval $c_n \stackrel{\text{def}}{=} \sqrt{\frac{\log(2Nn(n+1)\beta^{-1})}{2n}}$.

We now provide high confidence bounds on the number of times each sub-optimal node is visited.

**Theorem 3.** Assume $A^*$. Let $I$ denotes the set of nodes $i$ such that $\Delta_i > \delta_{d_i}$, where $d_i$ is the depth of node $i$. Define recursively the numbers $N_i$ associated to each node $i$ of a sub-optimal branch (i.e. for which $\Delta_i > 0$):

- If $i$ is a leaf, then $N_i \stackrel{\text{def}}{=} \frac{6\log(4N\beta^{-1}/\Delta_i^2)}{\Delta_i^2}$.

- It $i$ is not a leaf, then

$$N_i \stackrel{\text{def}}{=} \begin{cases} N_{i_1} + N_{i_2}, & \text{if } i \notin I \\ \min(N_{i_1} + N_{i_2}, \frac{6\log(4N\beta^{-1}/(\Delta_i - \delta_{d_i})^2)}{(\Delta_i - \delta_{d_i})^2}), & \text{if } i \in I \end{cases}$$

where $i_1$ and $i_2$ denote the children nodes of $i$. Then, with probability $1-\beta$, for all $n \geq 1$, for all sub-optimal nodes $i$, we have $n_i \leq N_i$.

*Proof.* We consider the event $\mathcal{E}$ under which $|X_{i,n} - \bar{X}_{i,n}| \leq c_n$ for all nodes $i$ and all times $n \geq 1$. Like in the previous result, the confidence interval $c_n = \sqrt{\frac{\log(2Nn(n+1)\beta^{-1})}{2n}}$ is chosen such that $\mathbb{P}(\mathcal{E}) \geq 1-\beta$.

Under $\mathcal{E}$, the smoothness assumption along an optimal path implies that for any optimal node $*$ in this branch, $B_{*,n_*}$ is a true upper bound on $\mu^*$, i.e. $\mu^* \leq B_{*,n_*}$.

Now, for any sub-optimal leaf $i$, using the same analysis as in Flat UCB, we deduce the bound $n_i \leq N_i$. Then, by backward induction on the depth, assume that $n_i \leq N_i$ for all sub-optimal nodes of depth $d + 1$. Let $i$ be a node of depth $d$. Then $n_i \leq n_{i_1} + n_{i_2} \leq N_{i_1} + N_{i_2}$. Now consider a sub-optimal node $i \in I$. If the node $i$ is chosen at round $n$, the form of the bound (7) implies that for any optimal node $*$, we have $B_{*,n_*} \leq B_{i,n_i}$. Since $\mu^* \leq B_{*,n_*}$ and $B_{i,n_i} \leq X_{i,n_i} + \delta_d + c_{n_i} \leq \mu_i + \delta_d + 2c_{n_i}$, we deduce $\mu^* \leq \mu_i + \delta_d + 2c_{n_i}$, which rewrites $\Delta_i - \delta_d \leq 2c_{n_i}$. Using the same argument as in the proof of Flat UCB, we deduce that for all $n \geq 1$, we have $n_i \leq \frac{6\log(4N\beta^{-1}/(\Delta_i-\delta_d)^2)}{(\Delta_i-\delta_d)^2}$. Thus $n_i \leq N_i$ at depth $d$, which finishes the inductive proof. □

Now we would like to compare the regret of BAST to that of Flat UCB. First, we expect a direct gain for nodes $i \in I$. Indeed, from the previous result, whenever a node $i$ of depth $d$ is such that $\Delta_i > \delta_d$, then this node will be visited, with high probability, at most $O(1/(\Delta_i - \delta_d)^2)$ times (neglecting log factors), independently of the number of leaves ($2^{D-d}$) in the branch starting from $i$. But we also expect an improved bound on $n_i$ whenever $\Delta_i > 0$ if at a certain depth $h \in [d, D]$, all nodes $j$ of depth $h$ in the branch $i$ satisfy $\Delta_j > \delta_h$.

The next result enables to further analyze the expected improvement over Flat UCB.

**Theorem 4.** *Let $\eta > 0$. Write $J_\eta \stackrel{\text{def}}{=} I_\eta \cap \mathcal{L}$ the set of $\eta$-optimal leaves. Assume $A_\eta$ with an exponential sequence $\delta_d = \delta\gamma^d$. Then, with probability at least $1 - \beta$, the pseudo regret satisfies, for all $n \geq 1$,*

$$\bar{R}_n \leq \sum_{i \in J_\eta, \Delta_i > 0} \frac{6}{\Delta_i} \log(\frac{4N}{\Delta_i^2 \beta}) + \frac{54(3\delta)^c}{\eta^{2+c}} \log(\frac{4N}{\eta^2 \beta})$$

$$\leq 6|J_\eta|\frac{1}{\Delta}\log(\frac{4N}{\Delta^2 \beta}) + \frac{54(3\delta)^c}{\eta^{2+c}}\log(\frac{4N}{\eta^2\beta}) \quad (8)$$

*where $c \stackrel{\text{def}}{=} \log(2)/\log(1/\gamma)$.*

The first term in the bound is the same as for Flat UCB, but the sum is performed only on the leaves $i \in J_\eta$ whose value is $\eta$-close to optimality. The second term is $O(1/\eta^{2+c})$ depends weakly (linearly) on the depth $D$ (through $\log(N)$). Thus, if $\eta$ is fixed and we increase the depth $D$ of the tree, the first term is dominant and we expect an important regret reduction compared to Flat UCB when the number $|J_\eta|$ of $\eta$-optimal leaves is small compared to the total number of leaves $2^D$.

Notice that this bound (8) has no explicit exponential dependency w.r.t. the depth $D$. However the $1/\Delta$ term is of order $O(1/\gamma^D)$ from our smoothness assumption.

*Proof.* We consider the same event $\mathcal{E}$ as in the proof of Theorem 3. Let $h$ be the smallest integer such that $\delta_h \leq \eta/3$. We have $h \leq \frac{\log(3\delta/\eta)}{\log(1/\gamma)} + 1$. Let $i$ be a node of depth $h$. If $i \in I_{2\eta/3}$, then, thanks to the assumption $A_\eta$, we have for all $j \in \mathcal{L}(i)$, $\mu_j \geq \mu_i - \eta/3 \geq \mu^* - \eta$, thus $j \in J_\eta$.

Now, if $i \notin I_{2\eta/3}$, then using similar arguments as in Theorem 3, the number of times $n_i$ this node is visited is at most $n_i \leq \frac{6\log(4N\beta^{-1}/(\Delta_i-\delta_h)^2)}{(\Delta_i-\delta_h)^2}$, but since $\Delta_i - \delta_h \geq 2\eta/3 - \eta/3 \geq \eta/3$, we have: $n_i \leq l/\eta^2$, writing $l = 54\log(4N\beta^{-1}/\eta^2)$.

Since there are $2^h$ nodes of depth $h$, the number of times that nodes of depth $h$ that do not belong to $I_{2\eta/3}$ are chosen is bounded by

$$2^h \frac{l}{\eta^2} \leq 2\left(\frac{3\delta}{\eta}\right)^{\frac{\log(2)}{\log(1/\gamma)}} \frac{l}{\eta^2} = \frac{2l(3\delta)^c}{\eta^{c+2}}$$

The pseudo regret is bounded by the sum for all $\eta$-optimal leaves $i \in J_\eta$ of $n_i \Delta_i$ (with $n_i \leq \frac{6}{\Delta_i^2}\log(\frac{4N\beta^{-1}}{\Delta_i^2})$ like in Theorem 2) plus the number of times nodes $i$ of depth $h$ such that $i \notin I_{2\eta/3}$ are chosen:

$$\bar{R}_n \leq \sum_{i \in J_\eta, \Delta_i > 0} \frac{6}{\Delta_i} \log(\frac{4N}{\Delta_i^2 \beta}) + \frac{54(3\delta)^c}{\eta^{2+c}} \log(\frac{4N}{\eta^2 \beta}).$$

which ends the proof. □

**Remark 1.** *Notice that if we choose $\delta = 0$, then BAST algorithm reduces to regular UCT (with a slightly different confidence interval), since in that case, the min in (7) gives the bounds $B_{i,n_i} = X_{i,n_i} + c_{n_i}$. Now if $\delta = \infty$, then the bounds is $B_{i,n_i} = \max_{j \in \mathcal{C}(i)} B_{j,n_j}$, which is simply Flat UCB. Thus BAST may be seen as a generic UCB-based bandit algorithm for tree search, that allows to take into account actual smoothness of the tree, if available.*

**Numerical experiments: global optimization of a noisy function.** We search the global optimum of an $[0,1]$-valued function, given noisy values. This is a continuum-armed bandit problem (see e.g. [3]). The domain $[0,1]$ is uniformly discretized into $2^D$ intervals $[\frac{j}{2^D}, \frac{j+1}{2^D}]_{0 \leq j < 2^D}$ (of center $y_j$), each one related to a leaf $j$ of a tree of depth $D$. The tree implements a recursive binary splitting of the domain, each node of depth $d$ corresponding to an interval of size $2^{-d}$. At time $t$, if the algorithm selects a leaf $j$, then the (binary) reward $x_t \stackrel{i.i.d.}{\sim} \mathcal{B}(f(y_j))$, a Bernoulli random variable with parameter $f(y_j)$ (i.e. $\mathbb{P}(x_t = 1) = f(y_j)$, $\mathbb{P}(x_t = 0) = 1 - f(y_j)$).

If we assume that $f$ is Lipschitz (with Lipschitz constant $L$), then the exponential smoothness assumption $\delta_d = \delta\gamma^d$ holds for all nodes with $\delta = L/2$ and $\gamma = 1/2$ (thus $c = 1$). In the experiments, we used the function $f(x) \stackrel{\text{def}}{=} \max\left(3.6x(1-x), 1 - \frac{1}{a}|1 - a - x|\right)$, where $a > 0$, which is plotted in Figure 2 for $a = 0.1$. Note that an immediate upper bound on the Lipschitz constant of $f$ is $L = 1/a$.

BAST concentrates its ressources on the good leaves: In Figure 2 we show the proportion $n_j/n$ of visits of each leaf $j$. We observe that, when $n$ increases the proportion of visits concentrates around the global maximum.

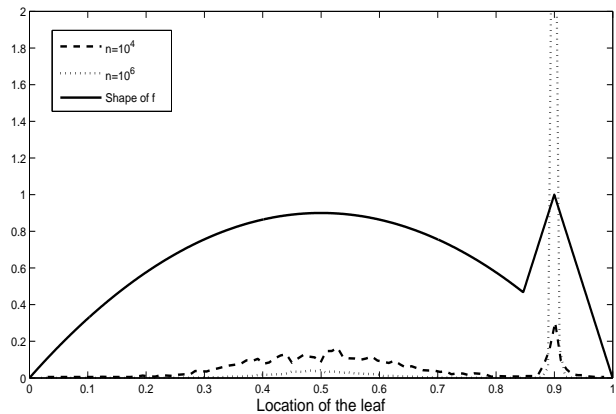

Figure 2: Function $f$ for $a = 0.1$ (plain curve) and (rescaled) proportion $n_j/n$ of leaves visitation for BAST with $\delta = 5$, $D = 10$ for $n = 10^4$ and $n = 10^6$.

We now compare the regret of Flat UCB, UCT and BAST algorithms for $a = 0.01$, which exhibits a function that possess a broad local optimal bump and a very picky global optimum. Figure 3 shows the cumulative regret per round $R_n/n$ (for $n = 10^7$) for BAST used with different values of the smoothness constant $\delta$. Starting from $\delta = 0$ (which corresponds to UCT with a slightly different confidence interval), the regret is poor. This holds also for regular UCT with the confidence interval defined by (1). This happens because the algorithm gets stuck in the local optimum, as illustrated in Section 2. When $\delta$ increases, the regret decreases (the global optimum of $f$ is reached) first, and then increases again, because the "cuts" performed by BAST are less and less frequent. For an infinite value of $\delta$ (which corresponds to Flat UCB) the regret is 0.23.

We observe that BAST with $\delta = L/2 = 1/(2a) = 50$ (which corresponds to the Lipschitz smoothness of $f$) outerforms both UCT and Flat UCB. The optimal performance is actually reached for a value $0 < \delta < 50$.

**Remark 2.** *If we know in advance that the function is locally smooth around its maximum, then one may use a smaller value of $\gamma$. For example, if the function is locally quadratic, then $\delta_d = \delta\gamma^d$ would hold for $\gamma = 1/4$ and appropriate $\delta$. This would cut more efficiently sub-optimal branches and yield improved performance (compared to $\gamma = 1/2$). Thus any a priori knowledge about the tree smoothness around optimality could be taken into account in the BAST bound (7).*

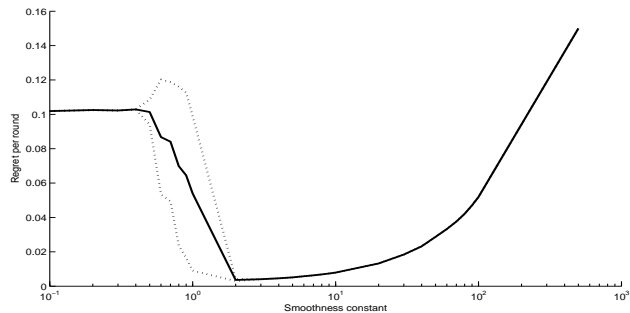

Figure 3: Cumulative regret per round $R_n/n$ for BAST with different values of the smoothness constant $\delta$. The dotted curves represent $+/-$ one standard deviation computed over 20 simulations. Here $a = 0.01$, $n = 10^7$, $D = 17$.

## 6  Growing trees

If the tree is too big (possibly infinite) to be represented, one may wish to discover it sequentially, exploring it at the same time as searching for an optimal value. We propose an incremental algorithm similar to the method described in [5] and [6]: The algorithm starts with only the root node. Then, at each stage $n$, it chooses which leaf, call it $i$, to expand next. Expanding a leaf means turning it into a node $i$, and adding in our current tree representation its children leaves $i_1$ and $i_2$, from which a reward (one for each child) is received. The process is then repeated in the new tree. Such an iterative growing tree requires a amount of memory $O(n)$ linear in the number of rounds.

Here, the tree is such that the value of each node $i$ is the maximum of the values of its children. Besides, the $A^*$ assumption says that from any optimal leaf $*$ of depth $d$, the received reward $x_n \in [0,1]$ at time $n$ is a random variable whose expected value satisfies: $\mu^* - \mathbb{E}[x_n|\mathcal{F}_{n-1}] \leq \delta_d$. Notice that we only make an assumption on the expectation of $x_n$, and not on its specific value at time $n$.

**Theorem 5.** *Assume $A^*$. Consider this incremental tree method using Algorithm 1 defined by the bound (7) with the confidence interval of a node $i$ of depth $d$:*

$$c_{d,n_i} \stackrel{\text{def}}{=} \sqrt{\frac{\log(2^{2d+1} n_i(n_i+1)\beta^{-1})}{2n_i}}.$$

*Then, with probability $1 - \beta$, for any sub-optimal node $i$ (of depth $d$), i.e. s.t. $\Delta_i > 0$, we have:*

$$n_i \leq \frac{6\log(2^{2d+2}\beta^{-1}/(\Delta_i - \delta_d)^2)}{(\Delta_i - \delta_d)^2}, \quad \text{if } \Delta_i > \delta_d,$$

$$\leq \frac{3}{2}\Big(\frac{\delta}{c}\Big)^c \Big(\frac{2+c}{\Delta_i}\Big)^{c+2} \log\Big(\frac{2^{2d}}{\beta}\frac{(2+c)^2}{\Delta_i^2}\Big)2^{-d},$$
$$\text{otherwise.}$$

Hence, any sub-optimal branch is visited a finite number of times. Thus, the number of trajectories that do not follow an optimal path up to a given depth is finite. Thus **this algorithm expands indefinitely the optimal branches only**.

*Proof.* Consider the event $\mathcal{E}$ under which $|X_{i,n_i} - \bar{X}_{i,n_i}| \leq c_{d,n_i}$ for all depths $d \geq 1$, for all nodes $i$ of depth $d$, for all times $n_i \geq 1$. Write $c_{d,n_i} = \sqrt{\frac{\log(2\beta_{d,n_i}^{-1})}{2n_i}}$ with $\beta_{d,n}^{-1} \stackrel{\text{def}}{=} 2^{2d}n(n+1)\beta^{-1}$. Since $\sum_{d\geq 1}\sum_{i=1}^{2^d}\sum_{n_i \geq 1}\beta_{d,n_i} = \beta$, we see that the confidence intervals $c_{d,n_i}$ are such that $\mathbb{P}(\mathcal{E}) \geq 1 - \beta$.

At round $n$, let $i$ be a node of depth $d$. If $\Delta_i > \delta_d$, then similarly to the proof of Theorem 3, we have $n_i \leq \frac{6\log(2^{2d+2}\beta^{-1}/(\Delta_i - \delta_d)^2)}{(\Delta_i - \delta_d)^2}$.

Otherwise, let $h$ be a depth such that $\delta_h < \Delta_i$ (thus $h > d$). This is satisfied for all integer $h \geq \frac{\log(\delta/\Delta_i)}{\log(1/\gamma)}$. Similarly, we deduce that the number of times $n_j$ a node $j$ of depth $h$ has been visited is bounded by $6\log(2^{2d+2}\beta^{-1}/(\Delta_i - \delta_h)^2)/(\Delta_i - \delta_h)^2$. Thus $i$ has been visited at most

$$n_i \leq \min_{h \geq \frac{\log(\delta/\Delta_i)}{\log(1/\gamma)}} 2^{h-d}\frac{6\log(2^{2d+2}\beta^{-1}/(\Delta_i - \delta_h)^2)}{(\Delta_i - \delta_h)^2}.$$

This function is minimized (neglecting the log term) for $h = \log(\frac{\delta(2+c)}{c\Delta_i})/\log(1/\gamma)$, which leads to the second bound. □

For illustration, Figure 4 shows the tree obtained when applied to the function optimization problem of previous section. This algorithm develops the tree in an asymmetric way, expanding in depth the optimal branch, leaving mainly unexplored sub-optimal branches.

## 7 Conclusion

BAST enables to take into account possible smoothness in the tree to perform efficient "cuts"[1] of sub-optimal branches, with high probability. Numerically, in all the problems we considered, BAST performed better than both UCT and Flat UCB for an appropriate smoothness coefficient $\delta \in (0, \infty)$. Experimental work seems to be necessary to obtain the best coefficient. However, if any additional smoothness of the optimal branch is provided, the smoothness sequence in the bound (7) may be refined, leading to improved performance. The use of variance estimate (see e.g.

---
[1]Note that this term may be misleading here since the UCB-based methods described here never explicitly delete branches

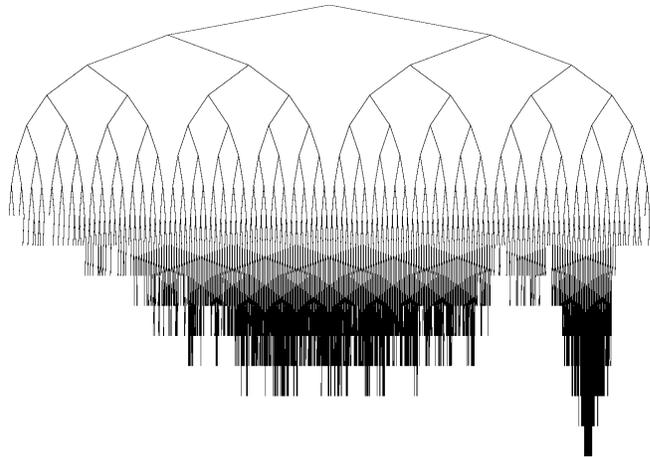

Figure 4: Tree resulting from the iterative growing BAST algorithm, after $n = 4000$ rounds, with $\delta = 5$, $a = 0.1$.

[1]) would tighten the confidence intervals and improve the performance too. However, it seems important to use true upper confidence bounds, in order to avoid bad cases as illustrated in regular UCT.

Extension to minimax search in 2 players, 0-sum games is possible and would yield similar results.

**Acknowledgements:** Warm thanks to Jean-François Hren for running all the numerical experiments.